\documentclass[conference]{IEEEtran}
\IEEEoverridecommandlockouts
\usepackage{cite}
\usepackage{amsmath,amssymb,amsfonts}
\usepackage{algorithmic}
\usepackage{graphicx}
\usepackage{textcomp}
\usepackage{xcolor}
\usepackage{multicol}
\def\BibTeX{{\rm B\kern-.05em{\sc i\kern-.025em b}\kern-.08em
    T\kern-.1667em\lower.7ex\hbox{E}\kern-.125emX}}

\usepackage{paralist}
\usepackage{xcolor}
\usepackage{amsmath}
\usepackage[linesnumbered, ruled]{algorithm2e}
\usepackage{hyperref}
\usepackage{url}
\usepackage{graphicx}
\usepackage{tabulary}
\usepackage{bookmark}
\usepackage{amsmath,amssymb,bm}
\usepackage{booktabs,floatrow}
\usepackage{multirow}
\usepackage{soul}
\usepackage{pgfplots}
\usetikzlibrary{pgfplots.groupplots}
\usepackage{tikz}
\usepgfplotslibrary{fillbetween}
\usetikzlibrary{arrows.meta}

\usepackage{colortbl}

\tikzset{%
  >={Latex[width=2mm,length=2mm]},
            base/.style = {rectangle, rounded corners, draw=black,
                          minimum width=1cm, minimum height=0.5cm,
                          text centered},
  activityStarts/.style = {base, minimum width=2cm, minimum height=1cm},
      startstop/.style = {base, fill=red!30},
    activityRuns/.style = {base, minimum width=2cm, minimum height=1cm, fill=blue!30},
         process/.style = {base,  draw=white, fill=white,
                          font=\ttfamily},
         l2rectangle/.style = {base},
}
\pgfkeys{%
  /tikz/on layer/.code={
    \pgfonlayer{#1}\begingroup
    \aftergroup\endpgfonlayer
    \aftergroup\endgroup
  }
}
\pgfplotsset{
    highlight/.code args={#1:#2}{
        \fill [every highlight] ({axis cs:#1,0}|-{rel axis cs:0,0}) rectangle ({axis cs:#2,0}|-{rel axis cs:0,1});
    },
    /tikz/every highlight/.style={
        on layer=\pgfkeysvalueof{/pgfplots/highlight layer},
        blue!20
    },
    /tikz/highlight style/.style={
        /tikz/every highlight/.append style=#1
    },
    highlight layer/.initial=axis background
}

\newcommand{\tad}{TadGAN}
\newcommand{\name}{TadGAN}
\newcommand\blfootnote[1]{%
  \begingroup
  \renewcommand\thefootnote{}\footnote{#1}%
  \addtocounter{footnote}{-1}%
  \endgroup
}

\definecolor{plotblue}{RGB}{0,96,179}
\definecolor{plotblack}{RGB}{0,64,119}
\newcommand{\tcr}{\textcolor[rgb]{1, 0, 0}}

\newcommand{\kanom}{``\textit{anomalies}''}

\definecolor{c0}{RGB}{181,72,60}
\definecolor{c1}{RGB}{210,103,82}
\definecolor{c2}{RGB}{223,142,112}
\definecolor{c3}{RGB}{231,180,156}
\definecolor{c4}{RGB}{242,221,207}

\definecolor{c5}{RGB}{222,232,243}
\definecolor{c6}{RGB}{195,213,231}
\definecolor{c7}{RGB}{154,190,214}
\definecolor{c8}{RGB}{112,157,196}
\definecolor{c9}{RGB}{74,125,177}

\begin{document}

\title{\name: Time Series Anomaly Detection Using Generative Adversarial Networks}



\makeatletter
\newcommand{\linebreakand}{%
  \end{@IEEEauthorhalign}
  \hfill\mbox{}\par
  \mbox{}\hfill\begin{@IEEEauthorhalign}
}
\makeatother

\author{\IEEEauthorblockN{Alexander Geiger\textsuperscript{*} }
\IEEEauthorblockA{\textit{MIT} \\
Cambridge, USA \\
geigera@mit.edu}
\and
\IEEEauthorblockN{Dongyu Liu\textsuperscript{*}}
\IEEEauthorblockA{\textit{MIT} \\
Cambridge, USA \\
dongyu@mit.edu}
\and
\IEEEauthorblockN{Sarah Alnegheimish}
\IEEEauthorblockA{\textit{MIT} \\
Cambridge, USA \\
smish@mit.edu}
\linebreakand 
\IEEEauthorblockN{Alfredo Cuesta-Infante}
\IEEEauthorblockA{\textit{Universidad Rey Juan Carlos} \\
Madrid, Spain\\
alfredo.cuesta@urjc.es}
\and
\IEEEauthorblockN{Kalyan Veeramachaneni}
\IEEEauthorblockA{\textit{MIT} \\
Cambridge, USA \\
kalyanv@mit.edu}
}

\maketitle

\begin{abstract}

Time series anomalies can offer information relevant to critical situations facing various fields, from finance and aerospace to the IT, security, and medical domains.
However, detecting anomalies in time series data is particularly challenging due to the vague definition of anomalies and said data's frequent lack of labels and highly complex temporal correlations. Current state-of-the-art unsupervised machine learning methods for anomaly detection suffer from scalability and portability issues, and may have high false positive rates.
In this paper, we propose~\name, an unsupervised anomaly detection approach built on Generative Adversarial Networks (GANs).
To capture the temporal correlations of time series distributions, we use LSTM Recurrent Neural Networks as base models for \textit{Generators} and \textit{Critics}.
\name~is trained with cycle consistency loss to allow for effective time-series data reconstruction.
We further propose several novel methods to compute reconstruction errors, as well as different approaches to combine reconstruction errors and \textit{Critic} outputs to compute anomaly scores.
To demonstrate the performance and generalizability of our approach, we test several anomaly scoring techniques and report the best-suited one.
We compare our approach to 8 baseline anomaly detection methods on 11 datasets from multiple reputable sources such as NASA, Yahoo, Numenta, Amazon, and Twitter.
The results show that our approach can effectively detect anomalies and outperform baseline methods in most cases (6 out of 11). Notably, our method has the highest averaged F1 score across all the datasets.
Our code is open source and is available as a \mbox{benchmarking tool}.

\end{abstract}



\blfootnote{* The two authors make equal contributions to this work. D. Liu and K. Veeramachaneni are the co-corresponding authors. Copyright: 978-1-7281-6251-5/20/\$31.00 ©2020 IEEE}

\section{Introduction}

The recent proliferation of temporal observation data has led to an increasing demand for time series anomaly detection in many domains, from energy and finance to healthcare and cloud computing. A time series anomaly is defined as a time point or period where a system behaves unusually~\cite{chandola2009anomaly}.
Broadly speaking, there are two types of anomalies: A \textit{point anomaly} is a single data point that has reached an unusual value, while a \textit{collective anomaly} is a continuous sequence of data points that are considered anomalous as a whole, even if the individual data points may not be unusual~\cite{chandola2009anomaly}.

Time series anomaly detection aims to isolate \textit{anomalous} subsequences of \textbf{varied lengths} within time series. One of the simplest detection techniques is \textit{thresholding}, which detects data points that exceed a normal range. However, many anomalies do not exceed any boundaries -- for example, they may have values that are purportedly ``normal,'' but are unusual at the specific time that they occur (i.e., \textit{contextual anomalies}). These anomalies are harder to identify because the context of a signal is often unclear \cite{ahmad2017unsupervised,chandola2009anomaly}.

\begin{figure*}[!t]
    \centering
    \includegraphics[width=\textwidth]{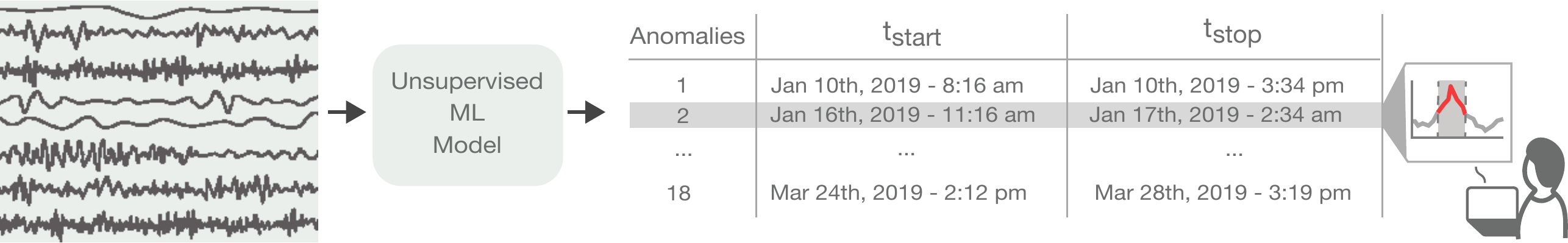}
    \caption{An illustration of time series anomaly detection using unsupervised learning. Given a multivariate time series, the goal is to find out a set of anomalous time segments that have unusual values and do not follow the expected temporal patterns.}
    \label{fig:anomaly_detection}
\end{figure*}

Various statistical methods have been proposed to improve upon thresholding, such as Statistical Process Control (SPC)~\cite{zheng2016self}, in which data points are identified as anomalies if they fail to pass statistical hypothesis testing. 
However, a large amount of human knowledge is still required to set prior assumptions on the models.
\begin{table}[!t]
\centering
\label{tab:wins}
\caption{The number of wins of a particular method compared with ARIMA, the traditional time series forecasting model, against an appropriate metric (f1 score) on 11 real datasets.}
\begin{tabular}{rc} 
\toprule
    \multicolumn{1}{c}{}  & Outperforms  \\ 
    \cline{2-2}
    Deep learning based method     & ARIMA, 1970 \cite{box2015time}  \\ 
    \hline
    LSTM AutoEncoder, 2016~\cite{malhotra2016lstm}   &   5   \\
    LSTM, 2018~\cite{Hundman2018}   &   5                   \\
    MAD-GAN, 2019~\cite{li2019mad}  &   0                   \\
    MS Azure, 2019~\cite{ren2019msftazure}  &   0           \\ 
    DeepAR, 2019~\cite{salinas2019deepar}  &   6            \\ 
    \hline
    TadGAN                & \textbf{8}                      \\
\bottomrule
\end{tabular}
\end{table}

Researchers have also studied a number of unsupervised machine learning-based approaches to anomaly detection. One popular method consists of segmenting a time series into subsequences (overlapping or otherwise) of a certain length and applying clustering algorithms to find outliers. Another learns a model that either predicts or reconstructs a time series signal, and makes a comparison between the real and the predicted or reconstructed values. High prediction or reconstruction errors suggest the presence of anomalies.

Deep learning methods~\cite{Kwon2017ASO} are extremely capable of handling non-linearity in complex temporal correlations, and have excellent learning ability. For this reason, they have been used in a number of time series anomaly detection methods ~\cite{Malhotra2015,Hundman2018,Goh2017AnomalyDI}, including tools created by companies such as Microsoft \cite{ren2019msftazure}. Generative Adversarial Networks (GANs)~\cite{Goodfellow2014} have also been shown to be very successful at generating time series sequences and outperforming state-of-the-art benchmarks~\cite{NIPS2019_8789}. Such a proliferation of methods invites the question: Do these new, complex approaches actually perform better than a simple baseline statistical method? To evaluate the new methods, we used 11 datasets (real and synthetic) that collectively have 492 signals and thousands of known anomalies to set-up a benchmarking system (see the details in Section~\ref{sec:experiments} and Table~\ref{tab:Baselines}). We implemented 5 of the most recent deep learning techniques introduced between 2016 and 2019, and compared their performances with that of a baseline method from the 1970s, \textsc{Arima}. While some methods were able to beat \textsc{Arima} on 50\% of the datasets, two methods failed to outperform it \mbox{at all (c.f. Table~\ref{tab:wins}). }

One of the foundational challenges of deep learning-based approaches is that their remarkable ability to fit data carries the risk that they could fit anomalous data as well. For example, autoencoders, using $L2$ objective function, can fit and reconstruct data extremely accurately - thus fitting the anomalies as well.
On the other hand, GANs may be ineffective at learning the generator to fully capture the data's hidden distribution, thus causing false alarms. Here, we mix the two methods, creating a more nuanced approach. Additionally, works in this domain frequently emphasize improving the deep learning model itself. However, as we show in this paper, improving post-processing steps could aid significantly in reducing the number of false positives.

In this work, we introduce a novel GAN architecture, \tad, for the time series domain. 
We use \tad~to reconstruct time series and assess errors in a contextual manner to identify anomalies.
We explore different ways to compute anomaly scores based on the outputs from \textit{Generators} and \textit{Critics}.
We benchmark our method against several well-known classical- and deep learning-based methods on eleven time series datasets. 
The detailed results can be found in Table~\ref{tab:Baselines}.

The key contributions of this paper are as follows:
\begin{compactitem}
  \item We propose a novel unsupervised GAN-reconstruction-based anomaly detection method for time series data. In particular, we introduce a cycle-consistent GAN architecture for time-series-to-time-series mapping.

   \item We identify two time series similarity measures suitable for evaluating the contextual similarity between original and GAN-reconstructed sequences. Our novel approach leverages GAN's \textit{Generator} and \textit{Critic} to compute robust anomaly scores at every time step. 

   \item We conduct an extensive evaluation using 11 time-series datasets from 3 reputable entities (NASA, Yahoo, and Numenta), demonstrating that our approach outperforms 8 other baselines. We further provide several insights into anomaly detection for time series data using GANs.
   
   \item  We develop a benchmarking system for time series anomaly detection. The system is open-sourced and can be extended with additional approaches and datasets\footnote{The software is available at github (\url{https://github.com/signals-dev/Orion})}.
   At the time of this writing, the benchmark includes 9 anomaly detection pipelines, 13 datasets, and 2 evaluation mechanisms.

\end{compactitem}


The rest of this paper is structured as follows.
We formally lay out the problem of time series anomaly detection in Section~\ref{sec:problem}.
Section~\ref{sec:related} presents an overview of the related literature.
Section~\ref{sec:formulation} introduces the details of our GAN model. 
We describe how to use GANs for anomaly detection in Section~\ref{sec:anomaly}, and evaluate our proposed framework in Section~\ref{sec:experiments}.
Finally, Section~\ref{sec:conclusion} summarizes the paper and reports our key findings.
\section{Unsupervised time series anomaly detection}
\label{sec:problem}
Given a time series $\mathbf{X}=(x^1, x^2, \cdots, x^T)$,  where $x^i \in \mathbf{R}^{M \times 1}$ indicates $M$ types of measurements at time step $i$, the goal of \textit{unsupervised} time series anomaly detection is to find a set of anomalous time segments $\bm{A_{seq}}=\{\bm{a_{seq}^1}, \bm{a_{seq}^2}, \cdots, \bm{a_{seq}^k}\}$, where $\bm{a_{seq}^i}$ is a continuous sequence of data points in time that show anomalous or unusual behaviors (Figure~\ref{fig:anomaly_detection}) -- values within the segment that appear not to comply with the expected temporal behavior of the signal. 
A few aspects of this problem make it both distinct from and more difficult than time series classification~\cite{fawaz2019deep} or supervised time series anomaly detection~\cite{qiu2019kpi}, as well as more pertinent to many industrial applications. We highlight them here:

\begin{itemize} 
\item[--]{No a priori knowledge of anomalies or possible anomalies}: Unlike with supervised time series anomaly detection, we do not have any previously identified ``known anomalies'' with which to train and optimize the model. Rather, we train the model to to learn the time series patterns, ask it to detect anomalies, and then check whether the detector identified anything relevant to \mbox{end users}.

\item[--]{Non availability of  ``normal baselines'' }: For many real-world systems, such as wind turbines and aircraft engines, simulation engines can produce a signal that resembles normal conditions, which can be tweaked for different control regimes or to account for degradation or aging. Such simulation engines are often physics-based and provide ``\textit{normal baselines},'' which can be used to train models such that any deviations from them are considered anomalous. Unsupervised time series anomaly detection strategies do not rely on the availability of such baselines, instead learning time series patterns from real-world signals -- signals that may themselves include anomalies or problematic patterns.

\item[--]{Not all detected anomalies are problematic}: Detected \kanom~ may not actually indicate problems, and could instead result from external phenomena (such as sudden shifts in environmental conditions), auxiliary information (such as the fact that a test run is being performed), or other variables that the algorithm did not consider, such as regime or control setting changes. Ultimately, it is up to the end user, the domain expert, to assess whether the anomalies identified by the model are problematic. Figure~\ref{fig:anomaly_detection} highlights how a trained unsupervised machine learning model can be used in real time for the \mbox{incoming data}. 

\item[--] {No clear segmentation possible}: Many signals, such as those associated with periodic time series, can be segmented -- for example, an electrocardiogram signal (ECG) can be separated into similar segments that pertain to periods~\cite{de2004automatic, qiu2019kpi}. The resulting segment clusters may reveal different collective patterns, along with anomalous patterns. We focus on signals that cannot be clearly segmented, making these approaches unfeasible. The length of $\bm{a^i}$ is also variable and is not known a priori, which further increases the difficulty. 

\item[--]{How do we evaluate these competing approaches?} For this, we rely on several datasets that contain ``known anomalies'', the details of which are introduced in Section~\ref{sec:datasets}. Presumably, the \kanom~are time segments that have been manually identified as such by some combination of algorithmic approaches and human expert annotation. These \kanom~are used to evaluate the efficacy of our proposed unsupervised models. More details about this can be found in Section~\ref{sec:metrics}.

\end{itemize}


\section{Related Work}
\label{sec:related}



Over the past several years, the rich variety of anomaly types, data types and application scenarios has spurred a range of anomaly detection approaches~\cite{hodge2004survey,chandola2009anomaly,Goldstein2016,habeeb2019real}.
In this section, we discuss some of the unsupervised approaches.
The simplest of these are out-of-limit methods, which flag regions where values exceed a certain threshold~\cite{Martinez-Heras2014,Decoste1997}. While these methods are intuitive, they are inflexible and incapable of detecting contextual anomalies. To overcome this, more advanced techniques have been proposed, namely proximity-based, prediction-based, and reconstruction-based anomaly detection (Table~\ref{tab:related_work}). 

\begin{table}[ht]
\centering
\begin{tabular}{@{}rp{4cm}@{}}
\toprule
\textbf{Methodology}  & \textbf{Papers} \\ \midrule
Proximity & 
\cite{breunig2000lof, angiulli2002fast, he2003discovering} \\
Prediction &
\cite{pena2013anomaly, torres2011detection, ahmad2017unsupervised, Hundman2018} \\
Reconstruction &
\cite{ringberg2007sensitivity, dai2013model, an2015variational, malhotra2016lstm} \\
Reconstruction (GANs) &
\cite{li2019mad, ijcai2019-616, NIPS2019_8789} \\ \bottomrule
\end{tabular}
\caption{Unsupervised approaches to time series anomaly detection.}
\label{tab:related_work}
\end{table}

\subsection{Anomaly Detection for Time Series Data.}




\textbf{Proximity-based methods} first use a distance measure to quantify similarities between objects -- single data points for point anomalies, or fixed length sequences of data points for collective anomalies. Objects that are distant from others are considered anomalies. This detection type can be further divided into distance-based methods, such as K-Nearest Neighbor (KNN)~\cite{angiulli2002fast} -- which use a given radius to define neighbors of an object, and the number of neighbors to determine an anomaly score -- and density-based methods, such as Local Outlier Factor (LOF)~\cite{breunig2000lof} and Clustering-Based Local Outlier Factor~\cite{he2003discovering}, which further consider the density of an object and that of its neighbors.
There are two major drawbacks to applying proximity-based methods in time series data: (1) a priori knowledge about anomaly duration and the number of anomalies is required; (2) these methods are unable to capture temporal correlations.

\textbf{Prediction-based methods}
learn a predictive model to fit the given time series data, and then use that model to predict future values. A data point is identified as an anomaly if the difference between its predicted input and the original input exceeds a certain threshold.
Statistical models, such as ARIMA~\cite{pena2013anomaly}, Holt-Winters~\cite{pena2013anomaly}, and FDA~\cite{torres2011detection}, can serve this purpose, but are sensitive to parameter selection, and often require strong assumptions and extensive domain knowledge about the data. 
Machine learning-based approaches attempt to overcome these limitations. \cite{ahmad2017unsupervised} introduce Hierarchical Temporal Memory (HTM), an unsupervised online sequence memory algorithm, to detect anomalies in streaming data.
HTM encodes the current input to a hidden state and predicts the next hidden state. A prediction error is measured by computing the difference between the current hidden state and the predicted hidden state.
Hundman et al.~\cite{Hundman2018} propose Long Short Term Recurrent Neural Networks (LSTM RNNs), to predict future time steps and flag large deviations from predictions. 


\textbf{Reconstruction-based methods} 
learn a model to capture the latent structure (low-dimensional representations) of the given time series data and then create a synthetic reconstruction of the data. Reconstruction-based methods assume that anomalies lose information when they are mapped to a lower dimensional space and thereby cannot be effectively reconstructed; thus, high reconstruction errors suggest a high chance of \mbox{being anomalous.}

Principal Component Analysis (PCA)~\cite{ringberg2007sensitivity}, a dimensionality-reduction technique, can be used to reconstruct data, but this is limited to linear reconstruction and requires data to be highly correlated and to follow a Gaussian distribution~\cite{dai2013model}.
More recently, deep learning based techniques have been investigated, including those that use Auto-Encoder (AE)~\cite{an2015variational}, Variational Auto-Encoder (VAE)~\cite{an2015variational} and LSTM Encoder-Decoder~\cite{malhotra2016lstm}. 

However, without proper regularization, these reconstruction-based methods can easily become overfitted, resulting in low performance. In this work, we propose the use of adversarial learning to allow for time series reconstruction. We introduce an intuitive approach for regularizing reconstruction errors.
The trained \textit{Generators} can be directly used to reconstruct more concise time series data -- thereby providing more accurate reconstruction errors -- while the \textit{Critics} can offer scores as a powerful complement to the reconstruction errors when computing an anomaly score.


\subsection{Anomaly Detection Using GANs.}

Generative adversarial networks can successfully perform many image-related tasks, including image generation~\cite{Goodfellow2014}, image translation~\cite{Zhu2017}, and video prediction~\cite{vondrick2016generating}, and researchers have recently demonstrated the effectiveness of GANs for anomaly detection in images~\cite{SCHLEGL201930,deecke2018anomaly}. 

\textbf{Adversarial learning for images.} Schlegl et al.~\cite{schlegl2017unsupervised} use the Critic network in a GAN to detect anomalies in medical images. They also attempt to use the reconstruction loss as an additional anomaly detection method, and find the inverse mapping from the data space to the latent space. This mapping is done in a separate step, after the GAN is trained. However, Zenati et al.~\cite{Zenati2018} indicate that this method has proven impractical for large data sets or real-time applications. They propose a bi-directional GAN for anomaly detection in tabular and image data sets, which allows for simultaneous training of the inverse mapping through an encoding network.

The idea of training both encoder and decoder networks was developed by Donahue et al.~\cite{Donahue2017} and Dumoulin et al.~\cite{Dumoulin2017}, who show how to achieve bidirectional GANs by trying to match joint distributions.
In an optimal situation, the joint distributions are the same, and the Encoder and Decoder must be inverses of each other. A cycle-consistent GAN was introduced by Zhu et al.~\cite{Zhu2017}, who have two networks try to map into opposite dimensions, such that samples can be mapped from one space to the other and vice versa.

\textbf{Adversarial learning for time series.} Prior GAN-related work has rarely involved time series data, because the complex temporal correlations within this type of data pose significant challenges to generative modeling. Three works published in 2019 are of note. First, to use GANs for anomaly detection in time series, Li et al.~\cite{li2019mad} propose using a vanilla GAN model to capture the distribution of a multivariate time series, and using the Critic to detect anomalies. Another approach in this line is BeatGAN~\cite{ijcai2019-616}, which is a Encoder and Decoder GAN architecture that allows for the use of the reconstruction error for anomaly detection in heartbeat signals. More recently, Yoon et al.~\cite{NIPS2019_8789} propose a time series GAN which adopts the same idea but introduces temporal embeddings to assist network training. However, their work is designed for time series representation learning instead of anomaly detection.

To the best of our knowledge, we are the first to introduce cycle-consistent GAN architectures for time series data, such that \textit{Generators} can be directly used for time series reconstructions. In addition, we systematically investigate how to utilize \textit{Critic} and \textit{Generator} outputs for anomaly score computation. A complete framework of time series anomaly detection is introduced to work with GANs.


\section{Adversarial Learning for time series reconstruction}
\label{sec:formulation}


The core idea behind reconstruction-based anomaly detection methods is to learn a model that can encode a data point (in our case, a segment of a time series) and then decode the encoded one (i.e., reconstructed one). An effective model should not be able to reconstruct anomalies as well as ``normal'' instances, 
because anomalies will lose information during encoding.
In our model, we learn two mapping functions between two domains $X$ and $Z$, namely $\mathcal{E}: X \to Z$ and $\mathcal{G}: Z \to X$ (Fig.~\ref{fig:model}). $X$ denotes the input data domain, describing the given training samples $\{(x_i^{1 \dots t})\}_{i=1}^N $, $x_i^{1 \dots t} \in X$. $Z$ represents the latent domain, where we sample random vectors $z$ to represent white noise. We follow a standard multivariate normal distribution, i.e., $z \sim \mathbb{P}_Z = \mathcal{N}(0, 1)$.
For notational convenience we use $x_i$ to denote a time sequence of length $t$ starting at time step $i$. 
With the mapping functions, we can reconstruct the input time series: ${x_i \to \mathcal{E}(x_i) \to \mathcal{G}(\mathcal{E}(x_i)) \approx \hat{x_i}}$.

We propose leveraging adversarial learning approaches to obtain the two mapping functions $\mathcal{E}$ and $\mathcal{G}$. As illustrated in Fig.~\ref{fig:model}, we view the two mapping functions as \textit{Generators}. 
Note that $\mathcal{E}$ is serving as an Encoder, which maps the time series sequences into the latent space, while $\mathcal{G}$ is serving as a Decoder, which transforms the latent space into the reconstructed time series.
We further introduce two adversarial \textit{Critics} (aka discriminators) $\mathcal{C}_x$ and $\mathcal{C}_z$. The goal of $\mathcal{C}_x$ is to distinguish between the real time series sequences from $X$ and the generated time series sequences from $\mathcal{G}(z)$, while $\mathcal{C}_z$ measures the performance of the mapping into latent space. 
In other words, $\mathcal{G}$ is trying to fool $\mathcal{C}_x$ by generating real-looking sequences.
Thus, our high-level objective consists of two terms: (1) \textit{Wasserstein losses}~\cite{Arjovsky}, to match the distribution of generated time series sequences to the data distribution in the target domain; and (2) \textit{cycle consistency losses}~\cite{Zhu2017}, to prevent the contradiction between $\mathcal{E}$ and  $\mathcal{G}$.





\subsection{Wasserstein Loss}
The original formulation of GAN that applies the standard adversarial losses (Eq.~\ref{eq:oriloss}) suffers from the mode \mbox{collapse problem.}
\begin{equation}
\label{eq:oriloss}
\mathcal{L} = \mathbb{E}_{x \sim \mathbb{P}_X} [\log\: \mathcal{C}_x(x)] + \mathbb{E}_{z \sim \mathbb{P}_Z} [\log(1 - \mathcal{C}_x(\mathcal{G}(z)))]    
\end{equation}
where $\mathcal{C}_x$ produces a probability score from 0 to 1 indicating the realness of the input time series sequence.
To be specific, the \textit{Generator} tends to learn a small fraction of the variability of the data, such that it cannot perfectly converge to the target distribution. This is mainly because the \textit{Generator} prefers to produce those samples that have already been found to be good at fooling the \textit{Critic}, and is reluctant to produce new ones, even though new ones might be helpful to capture other ``modes'' in the data.

To overcome this limitation, we apply Wasserstein loss~\cite{Arjovsky} as the adversarial loss to train the GAN. 
We make use of the Wasserstein-1 distance when training the \textit{Critic} network. Formally, let $\mathbb{P}_X$ be the distribution over $X$. For the mapping function $\mathcal{G}: Z \to X$ and its \textit{Critic} $C_x$, we have the \mbox{following objective}:
\begin{equation}
\min_{\mathcal{G}} \max_{\mathcal{C}_x \in \mathcal{\bold{C_x}}} V_{X}(\mathcal{C}_x, \mathcal{G})
\end{equation}
with 
\begin{equation}
\label{eq:Vx}
V_{X}(\mathcal{C}_x, \mathcal{G}) = \mathbb{E}_{x \sim \mathbb{P}_X} [\mathcal{C}_x(x)] - \mathbb{E}_{z \sim \mathbb{P}_Z} \left[\mathcal{C}_x(\mathcal{G}(z)))\right]    
\end{equation}
where $\mathcal{C}_x \in \mathcal{\bold{C_x}}$ which 
denotes the set of 1-Lipschitz continuous functions.
K-Lipschitz continuous functions are defined as follows:
$\|f(x_1)-f(x_2)\| \leq K\|x_1 - x_2\|, \forall x_1,x_2 \in dom\,f$.
The Lipschitz continuous functions constrain the upper bound of the function, further smoothing the function. Therefore, the weights will not change dramatically when updated with gradient descent methods. This reduces the risk of gradient explosion, and makes the model training more stable and reliable.
In addition, to enforce the 1-Lipschitz constraint during training, we apply a gradient penalty regularization term as introduced by Gulrajani et al.~\cite{Gulrajani}, which penalizes gradients not equal to 1 (cf. line 5).

\begin{figure}
\centering
\begin{tikzpicture}[node distance=1.3cm,
    every node/.style={fill=white, scale=0.8}, align=center]
  \node (dx)    [activityStarts]{$\mathcal{C}_x$};
  \node (x)     [process, below of=dx, left of=dx, xshift=-2cm]{$x \sim \mathbb{P}_X$};
  \node (l2)    [l2rectangle, right of=x, xshift=0.3cm]{$L2$};
  \node (gex)   [process, below of=dx]{$\mathcal{G}(\mathcal{E}(x))$};
  \node (gz)    [process, below of=dx, right of=dx, xshift=+2cm]{$\mathcal{G}(z)$};
  \node (e)     [activityRuns, below of=x, yshift=-0cm]{$\mathcal{E}$};
  \node (g)     [activityRuns, below of=gz, yshift=-0cm, xshift=-0.5cm]{$\mathcal{G}$};
  \node (ex)    [process, below of=e, yshift=-0cm]{$\mathcal{E}(x)$};
  \node (z)     [process, below of=g, yshift=-0cm, xshift=0.5cm]{$z \sim \mathbb{P}_Z$};   
  \node (dz)    [activityStarts, below of=ex, right of=ex, xshift=+2cm, yshift=0.2cm]{$\mathcal{C}_z$};

  \draw[->]     (x) |- (dx);
  \draw[->]     (gz) |- (dx);
  \draw         (x) -- (e);
  \draw[->]     (g.45) -- (gz);
  \draw[->]     (e) -- (ex);
  \draw         (z) -- (g.315);
  \draw[->]     (ex) |- (dz);
  \draw[->]     (z) |- (dz);
  \draw         (ex) -| (g.230);
  \draw[->]     (g.130) |- (gex);
  \draw[->]     (x) -- (l2);
  \draw[->]     (gex) -- (l2);

  \end{tikzpicture}
  
\caption{Model architecture: \textit{Generator} $\mathcal{E}$ is serving as an Encoder which maps the time series sequences into the latent space, while \textit{Generator} $\mathcal{G}$ is serving as a Decoder that transforms the latent space into the reconstructed time series. \textit{Critic} $\mathcal{C}_x$ is to distinguish between real time series sequences from $X$ and the generated time series sequences from $\mathcal{G}(z)$, whereas \textit{Critic} $\mathcal{C}_z$ measures the goodness of the mapping into the latent space. } \label{fig:model}
\end{figure}
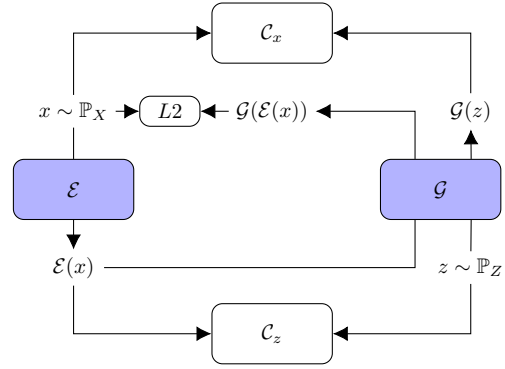

Following a similar approach, we introduce a Wasserstein loss for the mapping function $\mathcal{E}: X \to Z$ and its \textit{Critic} $C_z$. The objective is expressed as:
\begin{equation}
\label{eq:Vz}
\min_{\mathcal{E}} \max_{\mathcal{C}_z \in \mathcal{\bold{C_z}}} V_{Z}(\mathcal{C}_z, \mathcal{E})
\end{equation}
The purpose of the second \textit{Critic} $C_z$ is to distinguish between random latent samples $z \sim \mathbb{P}_Z$ and encoded samples $\mathcal{E}(x)$ with  $x \sim \mathbb{P}_X$. We present the model type and architecture for $\mathcal{E}$, $\mathcal{G}$, $\mathcal{C}_x$, $\mathcal{C}_z$ in section~\ref{setup}. 


\subsection{Cycle Consistency Loss}
The purpose of our GAN is to reconstruct the input time series: $x_i \to \mathcal{E}(x_i) \to \mathcal{G}(\mathcal{E}(x_i)) \approx \hat{x_i}$.
However, training the GAN with adversarial losses (i.e., Wasserstein losses) alone cannot guarantee mapping individual input $x_i$ to a desired output $z_i$ which will be further mapped back to $\hat{x}_i$. To reduce the possible mapping function search space, we adapt cycle consistency loss to time series reconstruction, which was first introduced by Zhu et al.~\cite{Zhu2017} for image translation tasks.
We train the generative network $\mathcal{E}$ and $\mathcal{G}$ with the adapted cycle consistency loss by minimizing the L2 norm of the difference between the original and the reconstructed samples:
\begin{equation}
\label{eq:VL2}
V_{L2}(\mathcal{E}, \mathcal{G}) = \mathbb{E}_{x \sim \mathbb{P}_X} [\lVert x - \mathcal{G}(\mathcal{E}(x)) \rVert_2]
\end{equation}
Considering that our target is anomaly detection, we use the L2 norm instead of L1 norm (the one used by Zhu et al.~\cite{Zhu2017} for image translation) to emphasize the impacts of anomalous values. In our preliminary experiments, we observed that adding the backward consistency loss (i.e., $\mathbb{E}_{z \sim \mathbb{P}_z} [\lVert z - \mathcal{E}(\mathcal{G}(z)) \rVert_2]$) did not improve performance.

\subsection{Full Objective}
Combining all of the objectives given in (\ref{eq:Vx}),(\ref{eq:Vz}) and (\ref{eq:VL2}) leads to the final MinMax problem:
\begin{equation}
\label{eq:MinMaxProblem}
    \min_{\{\mathcal{E}, \mathcal{G}\}} \max_{\{\mathcal{C}_x \in \mathcal{\bold{C_x}}, \mathcal{C}_z \in \mathcal{\bold{C_z}}\}} V_{X}(\mathcal{C}_x, \mathcal{G}) + V_{Z}(\mathcal{C}_z, \mathcal{E}) + V_{L2}(\mathcal{E}, \mathcal{G})
\end{equation}

The full architecture of our model can be seen in Figure \ref{fig:model}.
The benefits of this architecture with respect to anomaly detection are twofold. First, we have a \textit{Critic} $\mathcal{C}_x$ that is trained to distinguish between real and fake time series sequences, hence the score of the \textit{Critic} can directly serve as an anomaly measure. 
Second, the two \textit{Generators} trained with cycle consistency loss allow us to encode and decode a time series sequence. The difference between the original sequence and the decoded sequence can be used as a second anomaly detection measure.
For detailed training steps, please refer to the pseudo code (cf. line 1--14).
The following section will introduce the details of using \name~for anomaly detection.

\IncMargin{1.5em}
\begin{algorithm}[!t]
\SetKwInOut{Require}{Require}
\SetAlgoLined
\Indm
\Require{$m$, batch size.\\
 $epoch$, number of iterations over the data.\\
 $n_{critic}$, number of iterations of the critic per epoch.\\
 $\eta$, step size.\\
 }
 \Indp
 \For{each epoch}{
 \For{$\kappa = 0, \dots, n_{critic}$}{
      Sample $\{(x_i^{1\dots t})\}_{i=1}^m$ from real data.\\
      Sample $\{(z_i^{1\dots k})\}_{i=1}^m$ from random.\\
      $g_{w_{\mathcal{C}_x}} = \nabla_{w_{\mathcal{C}_x}} [\frac{1}{m} \sum_{i=1}^m \mathcal{C}_x(x_i) - \frac{1}{m} \sum_{i=1}^m \mathcal{C}_x(\mathcal{G}(z_i)) + gp(x_i, \mathcal{G}(z_i))]$\\
      $w_{\mathcal{C}_x} = w_{\mathcal{C}_x} + \eta \cdot \texttt{adam}(w_{\mathcal{C}_x}, g_{w_{\mathcal{C}_x}})$\\
      $g_{w_{\mathcal{C}_z}} = \nabla_{w_{\mathcal{C}_z}} [\frac{1}{m} \sum_{i=1}^m \mathcal{C}_z(z_i) - \frac{1}{m} \sum_{i=1}^m \mathcal{C}_z(\mathcal{E}(x_i)) + gp(z_i, \mathcal{E}(x_i))]$\\
      $w_{\mathcal{C}_z} = w_{\mathcal{C}_z} + \eta \cdot \texttt{adam}(w_{\mathcal{C}_z}, g_{w_{\mathcal{C}_z}})$\\
  }
  Sample $\{(x_i^{1\dots t})\}_{i=1}^m$ from real data.\\
  Sample $\{(z_i^{1\dots k})\}_{i=1}^m$ from random.\\
  $g_{w_{\mathcal{G},\mathcal{E}}} = \nabla_{w_\mathcal{G}, w_\mathcal{E}} [\frac{1}{m} \sum_{i=1}^m \mathcal{C}_x(x_i) - \frac{1}{m} \sum_{i=1}^m \mathcal{C}_x(\mathcal{G}(z_i)) + \frac{1}{m} \sum_{i=1}^m \mathcal{C}_z(z_i) - \frac{1}{m} \sum_{i=1}^m \mathcal{C}_z(\mathcal{E}(x_i)) + \frac{1}{m} \sum_{i=1}^m \|x_i - \mathcal{G}(\mathcal{E}(x_i))\|_2]$\\
  $w_{\mathcal{G}, \mathcal{E}},  = w_{\mathcal{G}, \mathcal{E}} + \eta \cdot \texttt{adam}(w_{\mathcal{G}, \mathcal{E}}, g_{w_{\mathcal{G}, \mathcal{E}}})$\\
 }
 $X = \{(x_i^{1\dots t})\}_{i=1}^n$\\
 \For{$i = 1, \dots, n$}{
    $\hat{x}_i = \mathcal{G}(\mathcal{E}(x_i))$\;
    $RE(x_i) = f(x_i, \hat{x}_i)$\;
    $score = \alpha Z_{RE}(x_i) + (1-\alpha) Z_{\mathcal{C}_x}(\hat{x}_i)$
 }
 \caption{TadGAN}
\end{algorithm}
\DecMargin{1.5em}

\section{Time-series GAN for anomaly detection (\name)}
\label{sec:anomaly}

Let us assume that the given time series is $\mathbf{X} =(x^1, x^2, \cdots, x^T)$,  where $x^i \in \mathbf{R}^{M \times 1}$ indicates $M$ types of measurements at time step $i$. For simplicity, we use $M=1$ in the later description. Therefore, $\mathbf{X}$ is now a univariate time series and $x^i$ is a scalar. The same steps can be applied for multivariate time series (i.e., when $M > 1$).

To obtain the training samples, we introduce a sliding window with window size $t$ and step size $s$ to divide the original time series into N sub-sequences $X = \{(x_i^{1 \dots t})\}_{i=1}^N$, where $N = \frac{T-t}{s}$. In practice, it is difficult to know the ground truth, and anomalous data points are rare. Hence, we assume all the training sample points are normal. In addition, we generate $Z = \{(z_i^{1 \dots k})\}_{i=1}^N$ from a random space following normal distribution, where $k$ denotes the dimension of the latent space. Then, we feed $X$ and $Z$ to our GAN model and train it with the objective defined in (\ref{eq:MinMaxProblem}). With the trained model, we are able to compute anomaly scores (or likelihoods) at every time step by leveraging the reconstruction error and \textit{Critic} output (cf. line~15--20).

\subsection{Estimating Anomaly Scores using Reconstruction Errors}

Given a sequence $x_i^{1 \dots t}$ of length $t$ (denoted as $x_i$ later), \name~ generates a reconstructed sequence of the same length: 
$x_i \to \mathcal{E}(x_i) \to \mathcal{G}(\mathcal{E}(x_i)) \approx \hat{x}_i$.
Therefore, for each time point $j$, we have a collection of reconstructed values $\{\hat{x}_{i}^{q}, i+q=j\}$
We take the median from the collection as the final reconstructed value $\hat{x}^j$. Note that in the preliminary experiments, we found that using the median achieved a better performance than using the mean. Now, the reconstructed time series is $(\hat{x}^1, \hat{x}^2, \cdots, \hat{x}^T)$. Here we propose three different types of functions (cf. line~18) for computing the reconstruction errors at each time step (assume the interval between neighboring time steps is the same).



\noindent \textbf{Point-wise difference.}\label{sec:ptDiff}
This is the most intuitive way to define the reconstruction error, which computes the difference between the true value and the reconstructed value at every time step:
\begin{equation}
\label{eq:PointDiff}
s_t = \left|x^t - \hat{x}^t \right|
\end{equation}

\noindent \textbf{Area difference.}\label{sec:areaDiff} This is applied over windows of a certain length to measure the similarity between local regions. It is defined as the average difference between the areas beneath two curves of length $l$:
\begin{equation}
\label{eq:AreaDiff}
s_t = \frac{1}{2*l} \left|\int_{t-l}^{t+l} x^t - \hat{x}^t \, dx\right|
\end{equation}
Although this seems intuitive, it is not often used in this context -- however, we will show in our experiments that this approach works well in many cases. Compared with the point-wise difference, the area difference is good at identifying the regions where small differences exist over a long period of time.
Since we are only given fixed samples of the functions, we use the trapezoidal rule to calculate the definite integral in the implementation.

\noindent \textbf{Dynamic time warping (DTW).}\label{dtw} DTW aims to calculate the optimal match between two given time sequences ~\cite{Bemdt1994} and is used to measure the similarity between local regions. We have two time series $X = (x_{t-1},x_{t-l+1},\dots,x_{t+l})$ and $\hat{X} = (\hat{x}_{t-1},\hat{x}_{t-l+1},\dots,\hat{x}_{t+l})$
and let $W\in\mathbf{R}^{2*l\times 2*l}$ be a matrix such that the $(i,j)$-th element is a distance measure between $x_i$ and $\hat{x}_j$, denoted as $w_k$.
We want to find the warp path $W^* = (w_1, w_2, \dots, w_K)$ that defines the minimum distance between the two curves, subject to boundary conditions at the start and end, as well as constraints on continuity and monotonicity.
The DTW distance between time series $X$ and $\hat{X}$ is defined as follows:
\begin{equation}
\label{eq:DTW}
s_t= W^*  = \textrm{DTW($X, \hat{X}$)} = \min_W \left[ \frac{1}{K} \sqrt{\sum_{k=1}^K w_k} \, \right] 
\end{equation}
Similar to area difference, DTW is able to identify the regions of small difference over a long period of time, but DTW can handle time shift issues as well.



\begin{center}
\begin{table*}[!t]
\caption{Dataset summary: overall the benchmark dataset contains a total of 492 signals and 2349 anomalies.}
\begin{tabulary}{2\textwidth}{lrrrrrrrrrrr}
\toprule
\multicolumn{1}{c}{} &
\multicolumn{2}{c}{NASA} & \multicolumn{4}{c}{Yahoo S5} & \multicolumn{5}{c}{NAB} \\
\cmidrule(r){2-3}\cmidrule(r){4-7}\cmidrule(r){8-12}
\textbf{Property}  &           SMAP &            MSL &  A1 &  A2 &  A3 &  A4 &  Art &  AdEx &  AWS &   Traf &     Tweets \\  
\midrule  
\textsc{\# signals}              &      53 &      27 &        67 &       100 &       100 &       100 &               6 &        5 &          17 &      7 &      10 \\
\textsc{\# anomalies}            &      67 &      36 &       178 &       200 &       939 &       835 &               6 &       11 &          30 &     14 &      33 \\  
\hspace{1em} point $(len=1)$&       0 &       0 &        68 &        33 &       935 &       833 &               0 &        0 &           0 &      0 &       0 \\  
\hspace{1em} collective $(len>1)$&      67 &      36 &       110 &       167 &         4 &         2 &               6 &       11 &          30 &     14 &      33 \\  
\textsc{\# anomaly points}    &   54696 &    7766 &      1669 &       466 &       943 &       837 &            2418 &      795 &        6312 &   1560 &   15651 \\ 
      
\hspace{1em} \# out-of-dist                  &      18126  &       642  &       861  &        153  &         21  &         49  &               123  &        15  &           210 &      86  &       520  \\  

\hspace{2em} (\% tot.)                  &   33.1\% &       8.3\% &       51.6\% &        32.8\% &         2.2\% &      5.9\% &                5.1\% &         1.9\% &            3.3\% &       5.5\% &        3.3\% \\  

\textsc{\# data points}          &  562800 &  132046 &     94866 &    142100 &    168000 &    168000 &           24192 &     7965 &       67644 &  15662 &  158511 \\
\textsc{is synthetic?}   &  &  &  & \checkmark & \checkmark & \checkmark & \checkmark &  &  &  & \\
\bottomrule
\vspace{-0.8cm}
\end{tabulary}
\label{tab:anomaly_summary}
\end{table*}
\end{center}

  


\subsection{Estimating Anomaly Scores with \textit{Critic} Outputs} 

During the training process, the \textit{Critic} $C_x$ has to distinguish between real input sequences and synthetic ones. Because we use the Wasserstein-1 distance when training $C_x$, the outputs can be seen as an indicator of how real (larger value) or fake (smaller value) a sequence is. Therefore, once the \textit{Critic} is trained, it can directly serve as an anomaly measure for time series sequences.  

Similar to the reconstruction errors, at time step $j$, we have a collection of \textit{Critic} scores $(c_{i}^{q}, i+q=j)$. We apply kernel density estimation (KDE) on the collection and then take the maximum value as the smoothed value $c^j$ 
Now the \textit{Critic} score sequence is $(c^1, c^2, \dots, c^T)$.
We show in our experiments that it is indeed the case that the \textit{Critic} assigns different scores to anomalous regions compared to normal regions. This allows for the use of thresholding techniques to identify the anomalous regions.

\subsection{Combining Both Scores}
\label{sec:anomalyscore}

The reconstruction errors $RE(x)$ and \textit{Critic} outputs $\mathcal{C}_x(x)$ cannot be directly used together as anomaly scores.
Intuitively, the larger $RE(x)$ and the smaller $\mathcal{C}_x(x)$ indicate higher anomaly scores. Therefore, we first compute the mean and standard deviation of $RE(x)$ and $\mathcal{C}_x(x)$, and then calculate their respective z-scores $Z_{RE}(x)$ and $Z_{\mathcal{C}_x}(x)$ to normalize both. Larger z-scores indicate high anomaly scores.

We have explored different ways to leverage $Z_{RE}(x)$ and $Z_{\mathcal{C}_x}(x)$.
As shown in Table~\ref{tab:evaluation} (row 1--4), we first tested three types of $Z_{RE}(x)$ and $Z_{\mathcal{C}_x}(x)$ individually.
We then explored two different ways to combine them (row 5 to the last row).
First, we attempt to merge them into a single value $\bm{a}(x)$ with a convex combination (cf. line~19)~\cite{li2019mad,schlegl2017unsupervised}: 
\begin{equation}
\bm{a}(x) = \alpha Z_{RE}(x) + (1-\alpha) Z_{\mathcal{C}_x}(x)     
\end{equation}
where $\alpha$ controls the relative importance of the two terms (by default $alpha=0.5$).
Second, we try to multiply both scores to emphasize the high values:
\begin{equation}
\bm{a}(x) = \alpha Z_{RE}(x) \odot Z_{\mathcal{C}_x}(x)
\end{equation}
where $\alpha=1$ by default.
Both methods result in robust anomaly scores. The results are reported in Section~\ref{sec:results}.

\subsection{Identifying Anomalous Sequences}
\label{sec:detecting}
\noindent \textbf{Finding anomalous sequences with locally adaptive thresholding:}
Once we obtain anomaly scores at every time step, thresholding techniques can be applied to identify anomalous sequences. 
We use sliding windows to compute thresholds, and empirically set the window size as $\frac{T}{3}$ and the step size as $\frac{T}{3*10}$.
This is helpful to identify contextual anomalies whose contextual information is usually unknown.
The sliding window size determines the number of historical anomaly scores to evaluate the current threshold.
For each sliding window, we use a simple static threshold defined as 4 standard deviations from the mean of the window. 
We can then identify those points whose anomaly score is larger than the threshold as anomalous.
Thus, continuous time points compose into anomalous sequences (or windows): $\{\bm{a}_{seq}^i, i=1,2,\dots,K \}$, where $\bm{a}_{seq}^i = (\bm{a}_{start(i)}, \dots, \bm{a}_{end(i)})$ . 

\noindent \textbf{Mitigating false positives:} 
The use of sliding windows can increase recall of anomalies but may also produce many false positives.
We employ an anomaly pruning approach inspired by Hundman et al.~\cite{Hundman2018} to mitigate false positives. At first, for each anomalous sequence, we use the maximum anomaly score to represent it, obtaining a set of maximum values $\{ \bm{a}_{max}^i, i=1,2,\dots,K \}$. Once these values are sorted in descending order, we can compute the decrease percent $p^i = (\bm{a}_{max}^{i-1} - \bm{a}_{max}^i) / \bm{a}_{max}^{i-1} $. When the first $p^i$ does not exceed a certain threshold $\theta$ (by default $\theta=0.1$), we reclassify all subsequent sequences (i.e., $\{\bm{a}_{seq}^j, i\leq j \leq K\}$) as normal.

\section{Experimental Results}

\label{sec:experiments}

\subsection{Datasets}
\label{sec:datasets}

To measure the performance of \tad, we evaluate it on multiple time series datasets. In total, we have collected 11 datasets (a total of 492 signals) across a variety of application domains. We use \textbf{spacecraft telemetry} signals provided by NASA\footnote{Spacecraft telemetry data: \url{https://s3-us-west-2.amazonaws.com/telemanom/data.zip}}, consisting of two datasets: Mars Science Laboratory (MSL) and Soil Moisture Active Passive (SMAP). In addition, we use \textbf{Yahoo S5} which contains four different sub-datasets \footnote{Yahoo S5 data can be requested here: \url{https://webscope.sandbox.yahoo.com/catalog.php?datatype=s&did=70}} The A1 dataset is based on real production traffic to Yahoo computing systems, while A2, A3 and A4 are all synthetic datasets. Lastly, we use \textbf{Numenta Anomaly Benchmark (NAB)}. NAB~\cite{lavin2015evaluating} includes multiple types of time series data from various application domains\footnote{NAB data: \url{https://github.com/numenta/NAB/tree/master/data}} We have picked five datasets: Art, AdEx, AWS, Traf, and Tweets.

Datasets from different sources contain different numbers of signals and anomalies, and locations of anomalies are known for each signal. Basic information for each dataset is summarized in Table~\ref{tab:anomaly_summary}. For each dataset, we present the total number of signals and the number of anomalies pertaining to them. We also observe whether the anomalies in the dataset are single ``point'' anomalies, or one or more collections. In order to suss out the ease of anomaly identification, we measure how out-of-the-ordinary each anomaly point is by categorizing it as ``\textit{out-of-dist}'' if it falls $4$ standard deviations away from the mean of all the data for a signal. As each dataset has some quality that make detecting its anomalies more challenging, this diverse selection will help us identify the effectiveness and limitations of each baseline.

\subsection{Experimental setup}\label{setup}

\subsubsection{Data preparation}

For each dataset, we first normalize the data betweeen $[-1,1]$. Then we find a proper interval over which to aggregate the data, such that we have several thousands of equally spaced points in time for each signal.
We then set a window size $t=100$ and step size $s=1$ to obtain training samples for \name.
Because many signals in the Yahoo datasets contain linear trends, we apply a simple detrending function (which subtracts the result of a linear least-squares fit to the signal) before training and testing.

\subsubsection{Architecture}
In our experiments, inputs to \name~are time series sequences of length 100 (domain $X$), and the latent space (domain $Z$) is 20-dimensional. 
We use a 1-layer bidirectional Long Short-Term Memory (LSTM) with 100 hidden units as \textit{Generator} $\mathcal{E}$, and a 2-layer bidirectional LSTM with 64 hidden units each as \textit{Generator} $\mathcal{G}$, where dropout is applied.
We add a 1-D convolutional layer for both \textit{Critics}, with the intention of capturing local temporal features that can determine how anomalous a sequence is. The model is trained on a specific signal from one dataset for 2000 iterations, with a batch size of 64.


\subsubsection{Evaluation metrics}
\label{sec:metrics}
We measure the performance of different methods using the commonly used metrics Precision, Recall and F1-Score.
In many real-world application scenarios, anomalies are rare and usually window-based (i.e. a continuous sequence of points---see Sec.~\ref{sec:detecting}). 
From the perspective of end-users, the best outcome is to receive timely true alarms without too many false positives (FPs), as these may waste time and resources.
To penalize high FPs and reward the timely true alarms, we present the following window-based rules:
(1) If a known anomalous window overlaps any predicted windows, a \textbf{TP} is recorded.
(2) If a known anomalous window does not overlap any predicted windows, a \textbf{FN} is recorded.
(3) If a predicted window does not overlap any labeled anomalous region, a \textbf{FP} is recorded.
This method is also used in Hundman et al's work~\cite{Hundman2018}.

\begin{center}
  \begin{table*}[!t]
  \caption{F1-Scores of baseline models using window-based rules. Color encodes the performance of the F1 score. One is evenly divided into 10 bins, with each bin associated with one color. From dark red to dark blue, F1 score increases from 0 to 1.}
  \begin{tabulary}{2\textwidth}{LLLLLLLLLLLLL}
  \toprule
    \multicolumn{1}{c}{} &
    \multicolumn{2}{c}{NASA} & \multicolumn{4}{c}{Yahoo S5} & \multicolumn{5}{c}{NAB} \\
    \cmidrule(r){2-3}\cmidrule(r){4-7}\cmidrule(r){8-12}
    \textbf{Baseline} 
    & MSL & SMAP
    & A1 & A2 & A3 & A4
    & Art & AdEx & AWS & Traf & Tweets
    & Mean\textpm SD\\
  \toprule
  \name  
  &{\cellcolor{c6}}\textbf{0.623} & {\cellcolor{c7}}\textbf{0.704}
  & {\cellcolor{c8}}\textbf{0.8} & {\cellcolor{c8}}0.867 & {\cellcolor{c6}}0.685 & {\cellcolor{c6}}0.6
  & {\cellcolor{c8}}\textbf{0.8} & {\cellcolor{c8}}\textbf{0.8} & {\cellcolor{c6}}0.644 & {\cellcolor{c4}}0.486 & {\cellcolor{c6}}\textbf{0.609}
  & \textbf{0.700\textpm 0.123}  \\
  \hline
  (P) LSTM   
  & {\cellcolor{c4}}0.46  & {\cellcolor{c6}}0.69  
  & {\cellcolor{c7}}0.744 & {\cellcolor{c9}}\textbf{0.98}  & {\cellcolor{c7}}0.772 & {\cellcolor{c6}}0.645
  & {\cellcolor{c3}}0.375 & {\cellcolor{c5}}0.538 & {\cellcolor{c4}}0.474 & {\cellcolor{c6}}\textbf{0.634} & {\cellcolor{c5}}0.543
  & 0.623\textpm 0.163\\
  (P) Arima   
  & {\cellcolor{c4}}0.492 & {\cellcolor{c4}}0.42  
  & {\cellcolor{c7}}0.726 & {\cellcolor{c8}}0.836 & {\cellcolor{c8}}\textbf{0.815} & {\cellcolor{c7}}\textbf{0.703} 
  & {\cellcolor{c3}}0.353 & {\cellcolor{c5}}0.583 & {\cellcolor{c5}}0.518 & {\cellcolor{c5}}0.571 & {\cellcolor{c5}}0.567
  & 0.599\textpm 0.148\\
  (C) DeepAR
  & {\cellcolor{c5}}0.583 & {\cellcolor{c4}}0.453
  & {\cellcolor{c5}}0.532 & {\cellcolor{c9}}0.929 & {\cellcolor{c4}}0.467 & {\cellcolor{c4}}0.454
  & {\cellcolor{c5}}0.545 & {\cellcolor{c6}}0.615 & {\cellcolor{c3}}0.39  & {\cellcolor{c6}}0.6 & {\cellcolor{c5}}0.542
  & 0.555\textpm 0.130\\
  
  (R) LSTM AE 
  & {\cellcolor{c5}}0.507 & {\cellcolor{c6}}0.672 
  & {\cellcolor{c6}}0.608 & {\cellcolor{c8}}0.871 & {\cellcolor{c2}}0.248 & {\cellcolor{c1}}0.163 
  & {\cellcolor{c5}}0.545 & {\cellcolor{c5}}0.571 & {\cellcolor{c7}}\textbf{0.764} & {\cellcolor{c5}}0.552 & {\cellcolor{c5}}0.542
  & 0.549\textpm 0.193\\
  
  (P) HTM      
  & {\cellcolor{c4}}0.412  & {\cellcolor{c5}}0.557 
  & {\cellcolor{c5}}0.588 & {\cellcolor{c6}}0.662 & {\cellcolor{c3}}0.325 & {\cellcolor{c2}}0.287  
  & {\cellcolor{c4}}0.455 & {\cellcolor{c5}}0.519 & {\cellcolor{c5}}0.571 & {\cellcolor{c4}}0.474 & {\cellcolor{c5}}0.526
  & 0.489\textpm 0.108\\

  (R) Dense AE 
  & {\cellcolor{c5}}0.507 & {\cellcolor{c7}}0.7  
  & {\cellcolor{c4}}0.472 & {\cellcolor{c2}}0.294 & {\cellcolor{c0}}0.074 & {\cellcolor{c0}}0.09  
  & {\cellcolor{c4}}0.444 & {\cellcolor{c2}}0.267 & {\cellcolor{c6}}0.64  & {\cellcolor{c3}}0.333 & {\cellcolor{c0}}0.057
  & 0.353\textpm 0.212\\
  
  (R) MAD-GAN   
  & {\cellcolor{c1}}0.111 & {\cellcolor{c1}}0.128  
  & {\cellcolor{c3}}0.37 & {\cellcolor{c4}}0.439 & {\cellcolor{c5}}0.589 & {\cellcolor{c4}}0.464 
  & {\cellcolor{c3}}0.324 & {\cellcolor{c2}}0.297 & {\cellcolor{c2}}0.273 & {\cellcolor{c4}}0.412 & {\cellcolor{c4}}0.444
  & 0.35\textpm 0.137\\
  (C) MS Azure  
  & {\cellcolor{c2}}0.218 & {\cellcolor{c1}}0.118 
  & {\cellcolor{c3}}0.352 & {\cellcolor{c6}}0.612 & {\cellcolor{c2}}0.257 & {\cellcolor{c2}}0.204 
  & {\cellcolor{c1}}0.125 & {\cellcolor{c0}}0.066 & {\cellcolor{c1}}0.173 & {\cellcolor{c1}}0.166 & {\cellcolor{c1}}0.118
  & 0.219\textpm 0.145\\
  
  \toprule
  \end{tabulary}
  \vspace{-0.5cm}
  \label{tab:Baselines}
  \end{table*}
  \vspace{-0.8cm}
\end{center}

\subsubsection{Baselines}
The baseline methods can be divided into three categories: prediction-based methods, reconstruction-based methods, and online commercial tools.

\noindent \textbf{ARIMA} (Prediction-based).
An autoregressive integrated moving average (ARIMA) model is a popular statistical analysis model that learns autocorrelations in the time series for future value prediction.
We use point-wise prediction errors as the anomaly scores to detect anomalies. 

\noindent \textbf{HTM} (Prediction-based). 
Hierarchial Temporal Memory (HTM)~\cite{ahmad2017unsupervised} has shown better performance over many statistical analysis models in the Numenta Anomaly Benchmark. It encodes the current input to a hidden state and predicts the next hidden state. Prediction errors are computed as the differences between the predicted state and the true state, which are then used as the anomaly scores for anomaly detection. 

\noindent \textbf{LSTM} (Prediction-based). 
The neural network used in our experiments consists of two LSTM layers with 80 units each, and a subsequent dense layer with one unit which predicts the value at the next time step (similar to the one used by Hundman et al.~\cite{Hundman2018}).
Point-wise prediction errors are used for anomaly detection.

\noindent \textbf{AutoEncoder} (Reconstruction-based).
Our approach can be viewed as a special instance of ``adversarial autoencoders''~\cite{makhzani2015adversarial}, $\mathcal{E\circ G:} \: X \to X$.
Thus, we compare our method with standard autoencoders with dense layers or LSTM layers~\cite{malhotra2016lstm}.
The dense autoencoder consists of three dense layers with 60, 20 and 60 units respectively. 
The LSTM autoencoder contains two LSTM layers, each with 60 units.
Again, a point-wise reconstruction error is used to detect anomalies.

\noindent \textbf{MAD-GAN} (Reconstruction-based).
This method~\cite{li2019mad} uses a vanilla GAN along with an optimal instance searching strategy in latent space to support multivariate time series reconstruction. We use MAD-GAN to compute the anomaly scores at every time step and then apply the same anomaly detection method introduced in Sec.~\ref{sec:detecting} to find anomalies.  

\noindent \textbf{Microsoft Azure Anomaly Detector} (Commercial tool).
Microsoft uses Spectral Residual Convolutional Neural Networks (SR-CNN) in which the models are applied serially~\cite{ren2019msftazure}. The SR model is responsible for saliency detection, and the CNN is responsible for learning a discriminating threshold. The output of the model is a sequence of binary labels that is attributed to each timestamp.

\noindent \textbf{Amazon DeepAR} (Commercial tool).
DeepAR is a probabilistic forecasting model with autoregressive recurrent networks~\cite{salinas2019deepar}. We use this model in a similar manner to LSTM in that it is a prediction-based approach. Anomaly scores are presented as the regression errors which are computed as the distance between the median of the predicted value and true value.

\begin{figure}[!t]
    \centering
    \includegraphics[width=\textwidth]{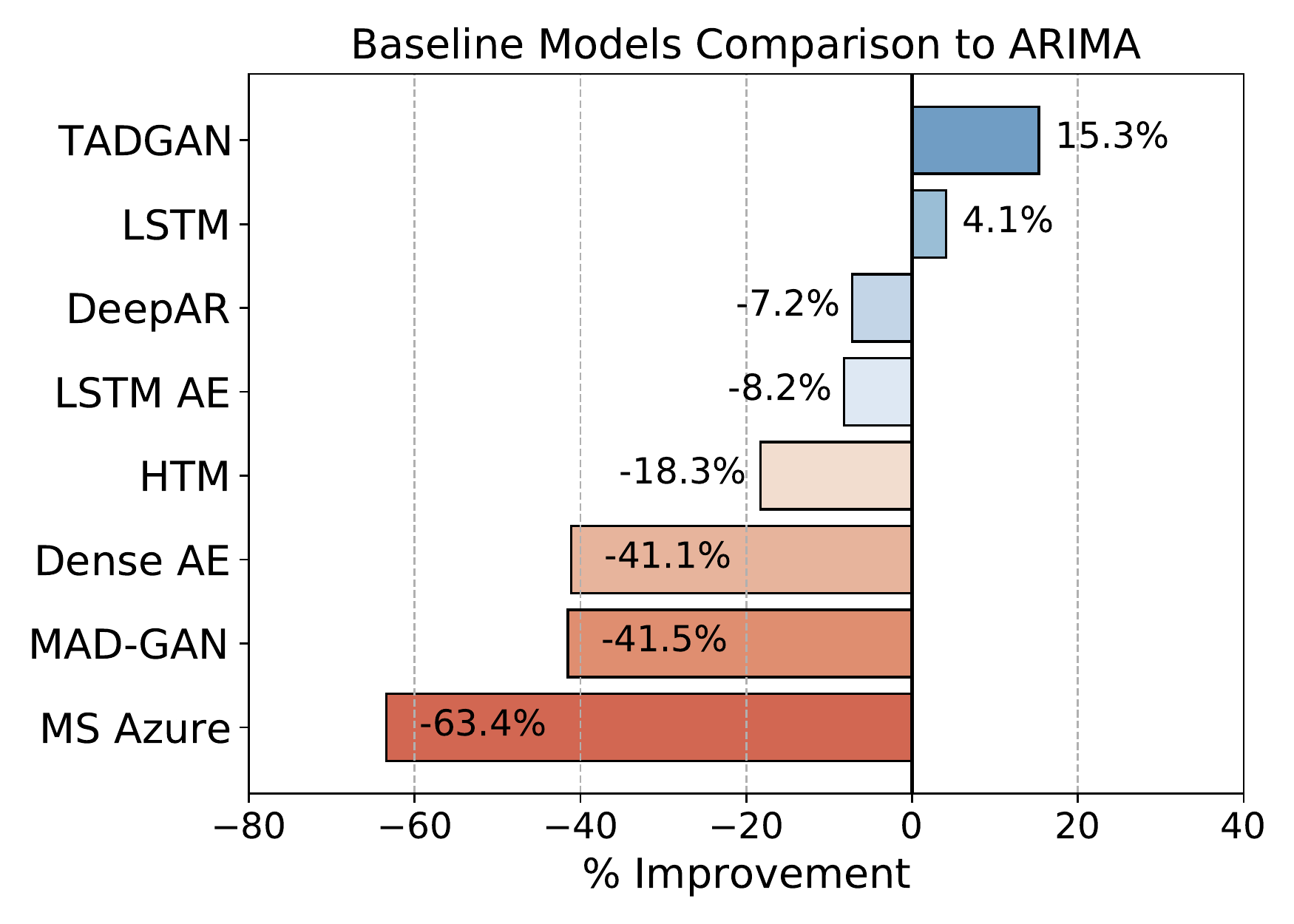}
    \caption{Comparing average F1-Scores of baseline models across all datasets to ARIMA. The x-axis represents the percentage of improvement over the ARIMA score by each one of the baseline models.}
    \label{fig:compare_to_arima}
\end{figure}

\subsection{Benchmarking Results}
\label{sec:results}
\textbf{\tad~outperformed all the baseline methods by having the highest averaged F1 score (0.7) across all the datasets.}
Table~\ref{tab:Baselines} ranks all the methods based on their averaged F1 scores (the last column) across the eleven datasets. 
The second (LSTM, 0.623) and the third (Arima, 0.599) best are both prediction-based methods and \tad~outperformed them by 12.46\% and 16.86\%, respectively, compared to the averaged F1 score.

\textbf{Baseline models in comparison to Arima.}
Figure~\ref{fig:compare_to_arima} depicts the performance of all baseline models with respect to Arima. It shows how much improvement in F1-Score is gained by each model. The F1-Score presented is the average across the eleven datasets. \tad~achieves the best overall improvement with an over 15\% improvement in score, followed by LSTM with a little over 4\% improvement. It's worth noting that all the remaining models struggle to beat Arima.

\textbf{Synthetic data v.s. real-world datasets.}
Although \tad~outperforms all baselines on average, we note that it ranks below Arima when detecting anomalies within synthetic dataset with point anomalies. Specifically, \tad~achieved an average of 0.717 while Arima scored an average of 0.784. However, \tad~still produces competitive results in both scenarios.

\begin{center}
  \begin{table*}[t]
   \caption{F1-Scores of all the variations of our model.}
  \begin{tabulary}{1.5\textwidth}{LLLLLLLLLLLLL}
  \toprule
   & \multicolumn{2}{c}{NASA} & \multicolumn{4}{c}{Yahoo S5} & \multicolumn{5}{c}{NAB}  \\
    \cmidrule(r){2-3}\cmidrule(r){4-7}\cmidrule(r){8-12}
  \textbf{Variation} 
  & MSL & SMAP 
  & A1 & A2 & A3 & A4 
  & Art & AdEx & AWS & Traf & Tweets
  & Mean+SD \\
  \toprule

  Critic       
  & {\cellcolor{c3}}0.393  & {\cellcolor{c6}}0.672
  & {\cellcolor{c2}}0.285  & {\cellcolor{c1}}0.118  & {\cellcolor{c0}}0.008  & {\cellcolor{c0}}0.024
  & {\cellcolor{c6}}0.625  & {\cellcolor{c0}}0      & {\cellcolor{c3}}0.35   & {\cellcolor{c1}}0.167 & {\cellcolor{c5}}0.548
  & 0.290\textpm 0.237\\
  Point        
  & {\cellcolor{c5}}0.585  & {\cellcolor{c5}}0.588 
  & {\cellcolor{c6}}0.674  & {\cellcolor{c7}}0.758 & {\cellcolor{c6}}0.628 & {\cellcolor{c5}}\textbf{0.6}
  & {\cellcolor{c5}}0.588  & {\cellcolor{c6}}0.611 & {\cellcolor{c5}}0.551 & {\cellcolor{c3}}0.383 & {\cellcolor{c5}}0.571
  & 0.594\textpm 0.086\\
  Area         
  & {\cellcolor{c5}}0.525 & {\cellcolor{c6}}0.655 
  & {\cellcolor{c6}}0.681  & {\cellcolor{c8}}0.82 & {\cellcolor{c5}}0.567 & {\cellcolor{c5}}0.523
  & {\cellcolor{c6}}0.625 & {\cellcolor{c6}}0.645 & {\cellcolor{c5}}0.59  & {\cellcolor{c4}}0.435 & {\cellcolor{c5}}0.559
  & 0.602\textpm 0.096\\
  DTW          
  & {\cellcolor{c5}}0.514 & {\cellcolor{c5}}0.581 
  & {\cellcolor{c6}}0.697  & {\cellcolor{c7}}0.794 & {\cellcolor{c6}}0.613 & {\cellcolor{c5}}0.547 
  & {\cellcolor{c7}}0.714  & {\cellcolor{c6}}0.69  & {\cellcolor{c6}}0.633 & {\cellcolor{c4}}0.455 & {\cellcolor{c5}}0.559
  & 0.618\textpm 0.095\\
  
  Critic$\times$Point 
  & {\cellcolor{c6}}0.619 & {\cellcolor{c6}}0.675 
  & {\cellcolor{c7}}0.703  & {\cellcolor{c7}}0.75  & {\cellcolor{c6}}\textbf{0.685} & {\cellcolor{c5}}0.536
  & {\cellcolor{c5}}0.588 & {\cellcolor{c5}}0.579  & {\cellcolor{c5}}0.576 & {\cellcolor{c4}}0.4 & {\cellcolor{c5}}0.59
  & 0.609\textpm 0.091\\
  Critic+Point 
  & {\cellcolor{c5}}0.529 & {\cellcolor{c6}}0.653
  & {\cellcolor{c8}}\textbf{0.8}  & {\cellcolor{c7}}0.78  & {\cellcolor{c5}}0.571  & {\cellcolor{c4}}0.44
  & {\cellcolor{c6}}0.625 & {\cellcolor{c5}}0.595  & {\cellcolor{c6}}\textbf{0.644} & {\cellcolor{c4}}0.439 & {\cellcolor{c5}}0.592
  & 0.606\textpm 0.111\\
  
  Critic$\times$Area  
  & {\cellcolor{c5}}0.578 & {\cellcolor{c7}}\textbf{0.704} 
  & {\cellcolor{c7}}0.719 & {\cellcolor{c8}}\textbf{0.867} & {\cellcolor{c5}}0.587 & {\cellcolor{c4}}0.46
  & {\cellcolor{c8}}\textbf{0.8}    & {\cellcolor{c6}}0.6  & {\cellcolor{c6}}0.6 & {\cellcolor{c4}}0.4 & {\cellcolor{c5}}0.571
  & 0.625\textpm 0.131\\
  Critic+Area  
  & {\cellcolor{c4}}0.493 & {\cellcolor{c6}}0.692
  & {\cellcolor{c7}}0.789 & {\cellcolor{c8}}0.847 & {\cellcolor{c4}}0.483 & {\cellcolor{c3}}0.367
  & {\cellcolor{c7}}0.75  & {\cellcolor{c7}}0.75  & {\cellcolor{c6}}0.607 & {\cellcolor{c4}}0.474 & {\cellcolor{c6}}0.6
  & 0.623\textpm 0.148\\
  
  Critic$\times$DTW   
  & {\cellcolor{c6}}\textbf{0.623} & {\cellcolor{c6}}0.68  
  & {\cellcolor{c6}}0.667 & {\cellcolor{c8}}0.82 & {\cellcolor{c6}}0.631 & {\cellcolor{c4}}0.497
  & {\cellcolor{c6}}0.667 & {\cellcolor{c6}}0.667 & {\cellcolor{c6}}0.61 & {\cellcolor{c4}}0.455 & {\cellcolor{c6}}0.605
  & \textbf{0.629\textpm 0.091}\\
  Critic+DTW   
  & {\cellcolor{c4}}0.462 & {\cellcolor{c6}}0.658  
  & {\cellcolor{c7}}0.735 & {\cellcolor{c8}}0.857 & {\cellcolor{c5}}0.523 & {\cellcolor{c3}}0.388
  & {\cellcolor{c6}}0.667 & {\cellcolor{c8}}\textbf{0.8}  & {\cellcolor{c6}}0.632 & {\cellcolor{c4}}\textbf{0.486} & {\cellcolor{c6}}\textbf{0.609}
  & 0.620\textpm 0.139\\
  \hline
  Mean
  & 0.532 & 0.655
  & 0.675 & 0.741 & 0.529 & 0.438
  & 0.664 & 0.593 & 0.579 & 0.409 & 0.580\\
  SD
  & 0.068 & 0.039 
  & 0.137 & 0.211 & 0.182 & 0.154
  & 0.067 & 0.209 & 0.081 & 0.087 & 0.02\\
  \toprule
  \end{tabulary}
  \label{tab:evaluation}
  \end{table*}
\end{center}

\textbf{How well do AutoEncoders perform?}
To view the superiority of GAN, we compare it to other reconstruction-based method such as LSTM AE, and Dense AE. One striking result is that the autoencoder alone does not perform well on point anomalies. We observe this as LSTM, AE, and Dense AE obtained an average F1 Score on A3 and A4 of 0.205 and 0.082 respectively, while \tad~and MAD-GAN achieved a higher score of 0.643 and 0.527 respectively. 
One potential reason could be that AutoEncoders are optimizing L2 function and strictly attempt to fit the data, resulting in that anomalies get fitted as well. However, adversarial learning does not have this type of issue.

\textbf{\tad~v.s. MadGAN.}
Overall, \tad~ (0.7) outperformed Mad-GAN (0.219) significantly. This fully demonstrates the usage of forward cycle-consistency loss (Eq.~\ref{eq:VL2}) which prevents the contradiction between two \textit{Generators} $\mathcal{E}$ and  $\mathcal{G}$ and paves the most direct way to the optimal $z_i$ that corresponds to the testing sample $x_i$.
Mad-GAN uses only vanilla GAN and does not include any regularization mechanisms to guarantee the mapping route $x_i \to z_i \to \hat{x}_i$. Their approach to finding the optimal $z_i$ is that they first sample a random $z$ from the latent space and then optimize it with the gradient descent algorithm by optimizing the anomaly detection loss.

\subsection{Ablation Study}


We evaluated multiple variations of \name, using different anomaly score computation methods for each (Sec.~\ref{sec:anomalyscore}).
The results are summarized in Table~\ref{tab:evaluation}.
Here we report some noteworthy insights.

\textbf{Using \textit{Critic} alone is unstable}, because it has the lowest average F1 score (0.29) and the highest standard deviation (0.237).
While only using \textit{Critic} can achieve a good performance in some datasets, such as SMAP and Art, its performance may also be unexpectedly bad, such as in A2, A3, A4, AdEx, and Traf.
No clear shared characteristics are identified among these five datasets (see Table~\ref{tab:anomaly_summary}). 
For example, some datasets contain only collective anomalies (Traf, AdEx), while other datasets, like A3 and A4, have point anomalies as the majority types.
One explanation could be that
\textit{Critic}'s behavior is unpredictable when confronted with anomalies ($x \nsim \mathbb{P}_X$), because it is only taught to distinguish real time segments ($x \sim \mathbb{P}_X$) from generated ones.

\textbf{DTW outperforms the other two reconstruction error types slightly.}
Among all variations, Critic$\times$DTW has the best score (0.629). Further, its standard deviation is smaller than most of the other variations except for Point, indicating that this combination is more stable than others. Therefore, this combination should be the safe choice when encountering new datasets without labels.



\textbf{Combining \textit{Critic} outputs and reconstruction errors does improve performance in most cases.}
In all datasets except A4, combinations achieve the best performance.
Let us take the MSL dataset as an example. We observe that when using DTW alone, the F1 score is $0.514$. Combining this with the \textit{Critic score}, we obtain a score of $0.623$, despite the fact that the F1 score when using \textit{Critic} alone is $0.393$.
In addition, we find that after combining the Critic scores, the averaged F1 score improves for each of the individual reconstruction error computation methods.
However, one interesting pattern is that for dataset A4, which consists mostly of point anomalies, using only point-wise errors achieve the best performance.

\textbf{Multiplication is a better option than convex combination. }
Multiplication consistently leads to a higher averaged F1 score than convex combination does when using the same reconstruction error type (e.g., \textit{Critic}$\times$Point v.s. \textit{Critic}+Point). Multiplication also has consistently smaller standard deviations. 
Thus, multiplication is the recommended way to combine reconstruction scores and \textit{Critic} scores. This can be explained by the fact that multiplication can better amplify high anomaly scores.



\subsection{Limitations and Discussion}

Here we compare our approach to one well-known GAN-based anomaly detection method~\cite{li2019mad}. However, there are many other GAN architectures tailored for time series reconstruction, such as Time-Series GAN~\cite{NIPS2019_8789}. Due to our modular design, any reconstruction-based algorithm of time series can employ our anomaly scoring method for time series anomaly detection. In the future, we plan to investigate various strategies for time series reconstruction and compare their performances to the current state-of-the-art.
Moreover, it is worth understanding how better signal reconstruction affects the performance of anomaly detection. In fact, it is expected that better reconstruction might overfit to anomalies. Therefore, further experiments are required to understand the relationship between reconstruction and detecting anomalies.



\section{Conclusion}
\label{sec:conclusion}
In this paper, we presented a novel framework, \name~, that allows for time series reconstruction and effective anomaly detection, showing how GANs can be effectively used for anomaly detection in time series data.
We explored point-wise and window-based methods to compute reconstruction errors. 
We further proposed two different ways to combine reconstruction errors and \textit{Critic} outputs to obtain anomaly scores at every time step.
We have also tested several anomaly-scoring techniques and reported the best-suited one in this work.
Our experimental results showed that 
(1) \name~outperformed all the baseline methods by having the highest averaged F1 score across all the datasets, and showed superior performance over baseline methods in 6 out of 11 datasets;
(2) window-based reconstruction errors outperformed the point-wise method; 
and (3) the combination of both reconstruction errors and critic outputs offers more robust anomaly scores, which help to reduce the number of false positives as well as increase the number of true positives.
Finally, our code is open source and is available as a tool for benchmarking time series datasets for anomaly detection.

\section{Acknowledgement}
The authors are grateful to SES S.A. of Betzdorf, Luxembourg, for their financial and non financial support in this work. 
Dr. Cuesta-Infante is funded by the Spanish Government research fundings RTI2018-098743-B-I00 (MICINN/FEDER) and Y2018/EMT-5062 (Comunidad de Madrid).
Alnegheimish is supported by King Abdulaziz City for Science and Technology (KACST).

\bibliographystyle{IEEEtran}
\bibliography{references}


\end{document}